\documentclass[twocolumn,a4paper,10pt]{article}

\usepackage[utf8]{inputenc}
\usepackage[T1]{fontenc}
\usepackage{lmodern}
\usepackage[margin=0.7in,columnsep=0.32in]{geometry}
\usepackage{graphicx}
\usepackage{booktabs}
\usepackage{caption}
\usepackage{subcaption}
\usepackage{amsmath,amssymb,amsfonts}
\usepackage{xcolor}
\usepackage{microtype}
\usepackage{authblk}
\usepackage{textcomp}
\usepackage{lineno}
\usepackage{setspace}
\usepackage{flushend}
\usepackage[hidelinks]{hyperref}
\usepackage[numbers,super,sort&compress,square]{natbib}
\usepackage{titlesec}
\titleformat{\section}{\normalfont\large\bfseries}{\thesection.}{0.5em}{}
\titleformat{\subsection}{\normalfont\normalsize\bfseries}{}{0em}{}
\titlespacing*{\section}{0pt}{1.4ex plus 0.6ex}{0.6ex plus 0.2ex}
\titlespacing*{\subsection}{0pt}{0.9ex plus 0.4ex}{0.3ex plus 0.1ex}
\title{\bfseries\large Fine-tuning a multimodal large language model for clinician-grade autism behavioral scoring from short home videos}

\author[1,$\dagger$]{Mohammadmahdi Honarmand}
\author[1,$\dagger$]{Parnian Azizian}
\author[2,3]{Aaron Kline}
\author[3]{Kae Nurge}
\author[4]{Zerin Nasrin Tumpa}
\author[2,3]{Saimourya Surabhi}
\author[2,3]{Kaitlyn Dunlap}
\author[4]{Yang Qian}
\author[4]{Ali Kargarandehkordi}
\author[5]{Sameer Neupane}
\author[5]{Peter Washington}
\author[2,3,*]{Dennis P. Wall}

\affil[1]{Department of Mechanical Engineering, Stanford University, Stanford, CA 94305, USA}
\affil[2]{Department of Pediatrics, Stanford University, Stanford, CA 94305, USA}
\affil[3]{Department of Biomedical Data Science, Stanford University, Stanford, CA 94305, USA}
\affil[4]{Department of Information and Computer Sciences, University of Hawaii at Manoa, Honolulu, HI 96822, USA}
\affil[5]{Division of Clinical Informatics and Digital Transformation, Department of Medicine, University of California San Francisco, San Francisco, CA 94158, USA}
\affil[*]{Correspondence: \href{mailto:dpwall@stanford.edu}{dpwall@stanford.edu}}
\affil[$\dagger$]{These authors contributed equally to this work.}

\date{}

\begin{document}

\twocolumn[
\begin{@twocolumnfalse}
\maketitle
\vspace{-2.5em}
\begin{abstract}
\noindent
Autism spectrum disorder (ASD) affects 1 in 31 US children, yet median age at diagnosis exceeds four years. Artificial intelligence pipelines that provide quantified diagnosis using easy to access observational data (e.g., home videos) could help with earlier diagnosis, and timely delivery of early treatments. We fine-tuned Gemini~2.5~Pro on 400 clinician-rated home videos with low-rank adaptation, training only on 30 behavioral features previously validated to produce reliable predictions when passed to various ML models. On 99 held-out children (49 ASD, 50 neurotypical), inter-rater reliability with clinicians (per-feature weighted Cohen's $\kappa$) improved by 40\% ($p<0.001$), with 27 of 28 evaluable features improving. As an emergent zero-shot capability, direct ASD diagnosis F1 improved by 53\% ($p<0.001$), matching or exceeding clinician outcomes. A Classifier-assisted pipelines using fine-tuned LLM-derived behavioral features matched clinician-scored inputs across all tested pathways and achieved 77\% accuracy (95\% CI: 68–85\%) and an AUC of 86\% (95\% CI: 78–92\%). Fine-tuned multimodal LLMs can serve as scalable behavioral feature extractors for use in autism assessment and diagnosis.
\end{abstract}
\vspace{1.5em}
\end{@twocolumnfalse}
]

\section{Introduction}

ASD affects 1 in 31 children in the United States, yet median age at diagnosis remains above four years,\citep{shaw2025prevalence} two or more years after the earliest reliable behavioral markers can be observed in naturalistic settings.\citep{pierce2019evaluation,mitroulaki2022first, perochon2023early} The cost of this delay is not academic: behavioral interventions delivered before age three yield the largest gains in cognitive, language, and adaptive outcomes,\citep{guthrie2023earlier, gabbay2022early} and every additional year before referral compresses the window in which those interventions are most effective.

The principal bottleneck is how the standard healthcare practice is set up. Commonly used tools such as the Vineland Adaptive Behavior Scale \cite{sparrow1989vineland} and the Autism Diagnostic Observation Schedule (ADOS)\citep{lord2000autism} require long evaluations, in person visits, and specialty-trained providers, an approach that does not scale for the population at risk in most regions of the world.\citep{durkin2015autism, mcconkey2022responding} Digital health solutions that can extract reliable behavioral information from accessible and enriched data types, e.g. short caregiver-uploaded home videos, could help achieve scale without loss of clinical accuracy. \cite{washington2023review, perochon2023early, tariq2018mobile, tariq2019detecting}

Machine learning approaches to home-video-based ASD diagnostic prediction have matured over the past decade. A small set of behavioral features tagged from short, naturalistic clips and fed to standard classifiers can discriminate ASD from typically developing children with accuracies above 90\% in retrospective and prospective cohorts.\citep{wall2012use,kosmicki2015searching,levy2017sparsifying,tariq2018mobile,leblanc2020feature} Subsequent work scaled these pipelines through privacy-preserving crowdsourcing,\citep{washington2021crowdsourced,washington2022crowd} added pose and gaze signals,\citep{kojovic2021using,yu2024video} ensembled multiple behavioral indicators for dynamic phenotyping,\citep{a18120764} and demonstrated clinical-grade performance from tablet-based digital phenotyping.\citep{perochon2023early} A meta-analysis confirmed the diagnostic value of home-video machine learning across independent cohorts,\citep{jin2024early} and an FDA-authorized device based on this paradigm has reached clinical practice.\citep{salomon2025analysis} Complementary purpose-built AI systems have also reported strong performance from controlled home-video protocols (AUROC~0.83 across 510 children)\citep{kim2025automated} and from multimodal mobile capture combining audio with diagnostic phenotyping tools (AUROC~0.94 for high-risk stratification across 1{,}242 children).\citep{bae2025multimodal} Alongside these performance gains, recent work has stressed uncertainty-aware deployment for such tools, modelling when a classifier should abstain or route a borderline case to human review.\citep{doi:10.1142/9789819824755_0009} Across this body of work, however, one bottleneck persists: producing the behavioral feature scores that power downstream classifiers still requires substantial human annotation effort.\citep{washington2021crowdsourced,chi2022classifying}

Although computer vision has matured at recognizing emotion and action from unconstrained, in-the-wild video\citep{qian2024advancinghumanactionrecognition,qian2023computervisionestimationemotion,Qian_2025_ICCV} and at test-time adaptation under distribution shift,\citep{honarmand2025fiestafisherinformationbasedefficient,Mutlu_2023_CVPR,Honarmand_2025_CVPR,Mutlu_2026_CVPR} the behavioral features are higher-order clinical judgments that conventional vision pipelines cannot be steered to produce, leaving the human-annotation bottleneck intact.

Multimodal large language models (LLMs) offer a route past this bottleneck because they ingest video frames, audio, and natural-language prompts in a single inference pass and can be steered by behavioral instruments expressed in plain language. We selected Gemini~2.5~Pro as the base model for this work because it is the leading publicly available multimodal LLM with native end-to-end video understanding, holds state-of-the-art results on long-form video benchmarks,\citep{team2024gemini,team2024gemini15,comanici2025gemini,fu2025video,li2024mvbench} and supports supervised fine-tuning through a managed cloud interface, which makes the resulting pipeline reproducible at the level of API calls. Supervised fine-tuning sits within a broader post-training and alignment toolkit that has become central to video foundation models.\citep{li2026video}. In medicine, domain-specific fine-tuning of multimodal LLMs has yielded substantial improvements over zero-shot baselines for radiology report generation,\citep{saab2024capabilities,yang2024advancing,tu2024towards} biomedical visual question answering,\citep{li2023llava,chen2024towards} surgical scene understanding,\citep{li2024llava,jin2024surgical,honarmand2024vidlpro} and parameter-efficient adaptation to specialised vision tasks.\citep{lian2024less,zhou2025ldp} Recent applications to autism include LLM-based deconstruction of clinical intuition,\citep{stanley2025large} scoresheet-aided diagnostic classification from clinical text,\citep{lin2025aiding} and language-disorder detection from spoken or written narratives.\citep{hu2025exploiting,hu2024exploiting}

For home-video-based ASD diagnostic prediction specifically, Azizian et al.\ recently evaluated seven Gemini variants as zero-shot substitutes for human raters on a 30-item behavioral instrument.\citep{azizian2025multimodal} They reported (i) a mean per-feature weighted Cohen's $\kappa$ of 0.40, below the $\kappa\geq0.60$ threshold conventionally associated with substantial clinical agreement, and (ii) low sensitivity (32.7\%) on the held-out global ASD-vs-neurotypical impression, which is the most clinically consequential single output. That work established the feasibility of multimodal LLMs as behavioral raters but stopped short of any model adaptation, leaving open whether a single, instructable foundation model can deliver clinician-grade behavioral scoring with modest amounts of labelled data.

Here we address that limitation directly. We fine-tuned Gemini~2.5~Pro on 400 clinician-rated home videos using low-rank adaptation (LoRA)\cite{hu2021lora} and evaluated the resulting model on a held-out test set of 99 children (49 ASD, 50 neurotypical) across five independent inference runs. The fine-tuning task was supervised only on the 30 behavioral features (Q1--Q30); the global ASD-vs-NT diagnostic label was held out from training. Four findings have direct implications for scalable autism screening. First, fine-tuning produces clinician-grade behavioral feature scores: mean per-feature weighted $\kappa$ rises from 0.40 to 0.56, with 27 of 28 evaluable features improving and the largest gains concentrated on imitation, joint attention, social affect, and sensory items, behaviors that are diagnostically central but acoustically or contextually subtle. Second, the held-out diagnostic label improves as an emergent zero-shot capability: fine-tuning nearly doubles sensitivity for the global ASD impression, raises accuracy by 18\%, and matches or exceeds clinician-scored inputs on accuracy and F1. Third, a classifier-assisted pipeline (a downstream supervised classifier trained on the fine-tuned LLM feature vector) matches clinician-scored inputs across three complementary metrics (accuracy, AUC, F1, chosen because together they capture threshold-free discrimination, calibration at the operating threshold, and class-imbalanced precision/recall). The strongest classifier, Random Forest, reaches 76.8\% accuracy, 85.9\% AUC, and 77.2\% F1 on fine-tuned features, statistically equivalent to the same Random Forest on clinician inputs (all $p>0.3$) and significantly above the base model ($p=0.021$ accuracy, $p=0.032$ AUC, $p=0.007$ F1). Fourth, classifier-assisted pipelines outperform direct LLM diagnosis on every metric tested. Together these results support the potential for a fine-tuned multimodal LLM as a practical, drop-in behavioral feature extractor for population-level diagnostic pipelines, while clarifying that diagnostic decisions are best left to a downstream supervised classifier rather than the LLM itself.

\section{Results}

\subsection{Cohort and study design}

The corpus comprised 595 short home videos of children, each annotated by trained clinicians on a 30-item feature set (Q1--Q30) and a separate global ASD-vs-NT diagnostic label.\citep{lord2000autism,azizian2025multimodal} Videos were drawn from three source families to capture phenotypic and demographic breadth: a US cohort centred on the GuessWhat study\citep{kalantarian2019guess} ($n=261$), curated public YouTube clips ($n=199$), and a Bangladeshi research study ($n=135$). Splits were stratified by source-family $\times$ diagnosis to preserve cross-split balance (training: $n=400$; validation: $n=96$; held-out test: $n=99$). All experiments report on the 99-video held-out test set (49 ASD, 50 neurotypical [NT]). ASD prevalence was matched across training (55.0\%) and validation (55.2\%); the test set is balanced (49.5\% ASD). Cohort composition by split is summarised in Table~\ref{tab:cohort}.

\begin{table*}[t]
\centering
\caption{\textbf{Cohort and dataset splits.} Counts are videos; values in parentheses give the diagnostic breakdown as ASD\,/\,NT\,/\,Other. The held-out test set is binary by design, so the test column reports ASD\,/\,NT only.}
\label{tab:cohort}
\small
\setlength{\tabcolsep}{6pt}
\begin{tabular}{lcccc}
\toprule
\textbf{Source family} & \textbf{Train} ($n{=}400$) & \textbf{Val} ($n{=}96$) & \textbf{Test} ($n{=}99$) & \textbf{Total} \\
                       & \footnotesize ASD/NT/Other & \footnotesize ASD/NT/Other & \footnotesize ASD/NT & \footnotesize ASD/NT/Other \\
\midrule
US cohort (GuessWhat)           & 171 (145/23/3)            & 41 (33/7/1)            & 49 (30/19)         & 261 (208/49/4)            \\
Public YouTube videos           & 137 (46/25/66)            & 33 (11/6/16)           & 29 (10/19)         & 199 (67/50/82)            \\
Bangladesh cohort               & 92 (29/15/48)             & 22 (9/3/10)            & 21 (9/12)          & 135 (47/30/58)            \\
\midrule
\textbf{Total}                  & \textbf{400 (220/63/117)} & \textbf{96 (53/16/27)} & \textbf{99 (49/50)} & \textbf{595 (322/129/144)} \\
\bottomrule
\end{tabular}
\end{table*}

\subsection{Fine-tuning improves per-feature behavioral agreement}

\begin{figure*}[t]
\centering
\includegraphics[width=0.95\linewidth]{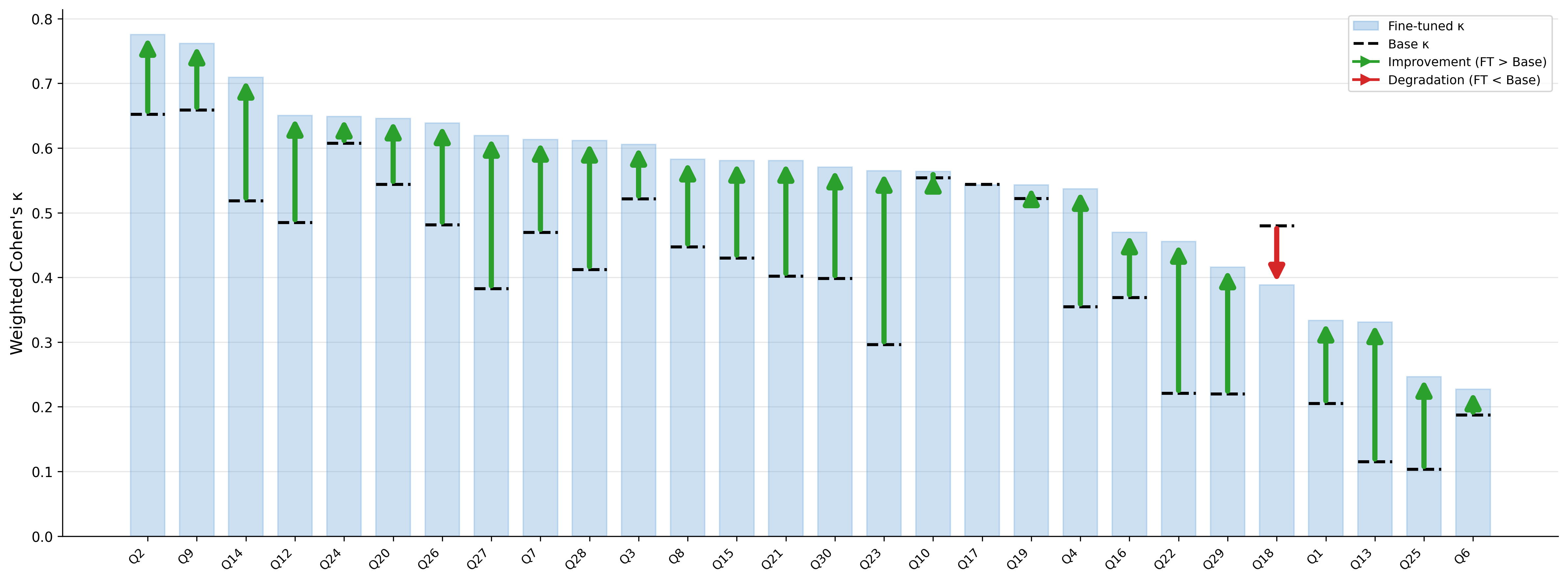}
\caption{\textbf{Fine-tuning improves per-feature agreement with clinicians on 27 of 28 evaluable behavioral features.} Bars denote weighted Cohen's $\kappa$ for the fine-tuned Gemini~2.5~Pro model on each of 28 behavioral features in the 99-video held-out test set; dashed black markers denote the matched base-model value. Green arrows mark features for which fine-tuning improved $\kappa$ (27/28); the single red arrow indicates head nod/shake, where $\kappa$ declined modestly. Items are sorted by post-fine-tuning $\kappa$.}
\label{fig:kappa}
\end{figure*}

Mean per-feature weighted Cohen's $\kappa$ rose from 0.40 (base model) to 0.56 (fine-tuned). Of the 28 features for which paired bootstrap confidence intervals could be computed, 27 improved after fine-tuning and one regressed (head nod/shake, $\Delta\kappa=-0.09$); two features (self-injurious behaviors and empathy) were not evaluable for paired comparison due to floor effects and insufficient base-model overlap, respectively (Fig.~\ref{fig:kappa}). The largest gains occurred on object play quality ($\Delta\kappa=+0.27$), enjoyment with others ($+0.24$), joint attention ($+0.23$), sensory sensitivity ($+0.22$), social initiation quality ($+0.20$), imitation ($+0.19$), and showing objects ($+0.18$). The highest absolute post-fine-tuning $\kappa$ values were observed on expressive language ($\kappa=0.78$, 95\% CI 0.69--0.85), responsiveness without name call ($\kappa=0.76$, 0.65--0.84), and imitation ($\kappa=0.71$, 0.60--0.79).

\subsection{Direct ASD diagnosis improves as an emergent zero-shot capability}

The global ASD-vs-NT diagnostic label was not used as a supervised target during fine-tuning: training optimized only the 30 feature set. Despite this, the fine-tuned model's diagnostic call improved substantially over the base model. Ensemble sensitivity rose from 32.7\% (16/49 ASD children correctly identified; 95\% CI 0.21--0.47) at base to 63.3\% (31/49; 95\% CI 0.49--0.76) after fine-tuning, a $+30.6$ percentage-point gain. Overall accuracy rose from 60.6\% to 71.7\%, F1 from 0.45 to 0.69 ($p<0.001$ by paired bootstrap), with specificity remaining at 80\% (Fig.~\ref{fig:clf}, right panel). Per-run sensitivity ranged from 0.58 to 0.65 across the five inference runs. The fine-tuned model also matched or exceeded clinician-scored direct diagnosis on accuracy (0.717 vs 0.697) and F1 (0.69 vs 0.63) by point estimates, with no significant clinician-vs-fine-tuned difference ($p=0.732$ accuracy, $p=0.483$ F1). In absolute terms the base model missed 33 of 49 ASD children, while the fine-tuned model missed 18, at the cost of four additional false-positive referrals (6 to 10 of 50 NT children).

\subsection{Classifier-assisted pipelines match clinician-scored inputs}

\begin{figure*}[t]
\centering
\includegraphics[width=0.95\linewidth]{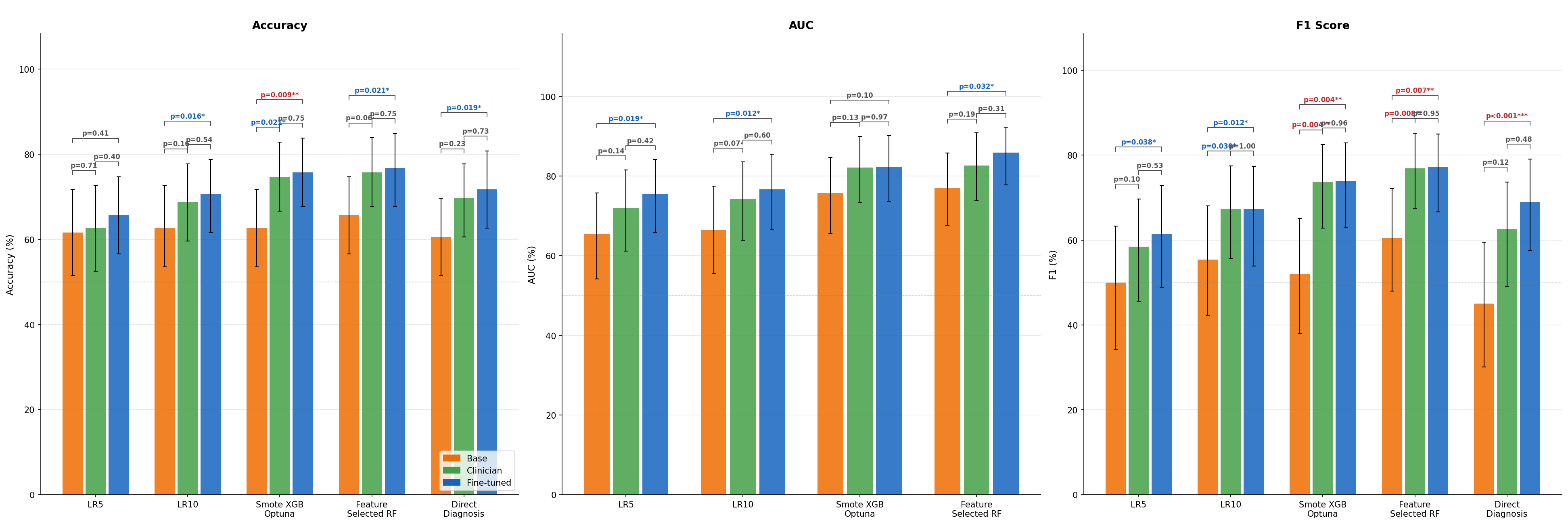}
\caption{\textbf{Fine-tuned LLM features close the gap with clinician-scored features and significantly outperform the base model across all five diagnostic pathways.} Each panel shows accuracy, AUC, and F1 for the diagnostic pathways under three input conditions: base Gemini~2.5~Pro (orange), clinician annotations (green), and fine-tuned Gemini~2.5~Pro (blue). Error bars are 95\% bootstrap confidence intervals; bracket annotations report two-sided paired bootstrap $p$-values. Across every clinician-vs-fine-tuned comparison the gap is non-significant ($p>0.3$); across base-vs-fine-tuned comparisons the fine-tuned model wins, with statistical significance for the headline metrics (full p-values reported in text).}
\label{fig:clf}
\end{figure*}

We next evaluated whether our fine-tuned LLM feature measurements can match clinician inputs that subsequently get submitted validated downstream classifiers. Five diagnostic pathways were tested on the 99-video held-out test set: two non-linear classifiers (Random Forest and XGBoost); two pre-trained logistic-regression classifiers from previously validated home-video pipelines\citep{tariq2018mobile,kosmicki2015searching,levy2017sparsifying} (LR10 on ten ADOS Module~3 features for verbally fluent children, LR5 on five ADOS Module~2 features for children with limited speech); and direct LLM diagnosis. Each pathway was scored under three input conditions: base Gemini, clinician annotations, and fine-tuned Gemini. Three metrics were tracked because together they summarise model behaviour comprehensively: AUC (threshold-free discrimination), accuracy (calibration at the operating threshold), and F1 (precision-recall balance under class imbalance). Pairwise comparisons used paired bootstrap tests with 10{,}000 resamples (Fig.~\ref{fig:clf}).

Across every pairing of condition and metric, fine-tuned features matched clinician-scored features: all 14 clinician-vs-fine-tuned comparisons were non-significant ($p\geq0.31$; all $p\geq0.40$ excluding the Random Forest AUC), and 12 of 14 base-vs-fine-tuned comparisons reached statistical significance in favour of the fine-tuned model.

The strongest classifier was Random Forest. On fine-tuned features it reached 76.8\% accuracy (95\% CI 67.7--84.8), 85.9\% AUC (77.7--92.2), and 77.2\% F1 (66.7--85.0), matching the same Random Forest on clinician inputs (clinician-vs-fine-tuned: $p=0.746$ accuracy, $p=0.306$ AUC, $p=0.950$ F1) and significantly outperforming the base-model Random Forest (base-vs-fine-tuned: $p=0.021$ accuracy, $p=0.032$ AUC, $p=0.007$ F1). XGBoost showed the same pattern on fine-tuned features (75.8\% accuracy [67.7--83.8], 82.2\% AUC [73.6--90.1], 0.74 F1 [0.63--0.83]; base-vs-fine-tuned $p=0.009$/0.10/0.004). LR10 and LR5 also followed the pattern (LR10 AUC 0.77 [0.67--0.85] vs 0.66 base, $p=0.012$; LR5 AUC 0.75 [0.66--0.84] vs 0.66 base, $p=0.019$).

\subsection{Classifier-assisted pipelines outperform direct LLM diagnosis}

Within the fine-tuned condition, all four classifier-assisted pathways outperformed direct LLM diagnosis on AUC and F1. The strongest performance came from the non-linear classifiers (Random Forest AUC 0.86, F1 0.77; XGBoost AUC 0.82, F1 0.74), with the linear classifiers also showing substantial gains (LR10 AUC 0.77, F1 0.67); direct diagnosis F1 was 0.69 with no AUC because the LLM emits a single binary call. The largest base-vs-fine-tuned effect was on direct-diagnosis F1 ($+0.24$, $p<0.001$), and the highest absolute fine-tuned performance was on the Random Forest classifier-assisted pathway.


\section{Discussion}

One major finding from our work is a substantial improvement in inter-rater reliability between the multimodal LLM and trained clinicians: mean per-feature weighted Cohen's $\kappa$ rose by 40\% with fine-tuning, with 27 of 28 evaluable features improving and the largest gains concentrated on diagnostically central items including object play quality, joint attention, social initiation, imitation, and social affect. This places per-feature reliability within the range typical of trained human raters, demonstrating that supervised fine-tuning on relatively modest clinician-labeled data (400 videos) can recover most of the agreement deficit that prior zero-shot LLMs left unresolved. As an emergent zero-shot capability---the global diagnostic label was held out from training---direct ASD diagnosis F1 also improved by 53\% ($p<0.001$), with the fine-tuned model matching or exceeding clinician-scored direct diagnosis. When the LLM-derived feature scores are passed to a validated downstream classifier, the resulting pipeline matches clinician-scored inputs across every diagnostic pathway tested (all clinician-vs-fine-tuned $p>0.3$) while significantly outperforming the base model. For a triage layer intended to flag children for follow-up evaluation, the fine-tuned model substantially reduces missed cases relative to the base model, a clinically appropriate trade.

The pattern of per-feature improvement is consistent with this clinical reading. The features that benefit most from fine-tuning, including object play quality, joint attention, social initiation, and imitation, are diagnostically central items where the relevant signal is observable but requires contextual integration of multiple cues across a 60--90~second clip. Fine-tuning aligns these contextual judgements with the clinician's scoring conventions rather than the model's general prior about typical child behavior. Items that remain difficult after fine-tuning, such as self-injurious behaviors and stereotyped language, share a structural property: the underlying behavior is rare or acoustically subtle in brief naturalistic clips, so the limiting factor is the observation window rather than the model. Longer clips, structured elicitation protocols, and supplementary parent-reported information are natural engineering responses.

The most striking finding is that the global ASD diagnostic call improved despite never being a supervised target. Training optimised only the 30 feature set, yet the held-out diagnostic label gained more than 30 percentage points of sensitivity and 24 points of F1. This emergent zero-shot transfer indicates that supervised feature-level alignment reshapes the model's underlying behavioral representation rather than merely improving local item calibration: by learning to score the constituent behaviors as a clinician would, the model implicitly acquires the integrative reasoning needed to translate those behaviors into a diagnostic impression, even though no diagnostic supervision was ever provided. This is consistent with broader observations that supervised fine-tuning on structured intermediate targets can elicit downstream capabilities that were not directly trained.\citep{saab2024capabilities,yang2024advancing,lin2025aiding}

Why does the structured classifier-assisted pipeline (LLM features into a supervised classifier) match or beat direct LLM diagnosis in every comparison? When asked to make a diagnostic call directly, the model must integrate multiple heterogeneous behavioral signals in a single autoregressive generation step, with no intermediate representation that training can supervise and no downstream mechanism to reweight features by their diagnostic informativeness in the relevant population. Forcing the model to score 30 behavioral features first changes the computation: each feature score is a supervised target with clinician labels, and the downstream classifier learns which combination of features is diagnostically predictive in a structured, interpretable, and cross-validatable way, mirroring how clinicians integrate observation into impression.\citep{lord2000autism,washington2023review} This decomposition principle has direct implications for the deployment of multimodal LLMs across other clinical observational instruments such as behavioral, developmental, and psychiatric domains that share an item-level scoring structure.

A realistic deployment scenario would combine the elements identified here. A caregiver records a 1--3~minute home video through a digital health platform. The fine-tuned LLM scores 30 behavioral features across five inference runs and the binary diagnostic call is aggregated by majority vote, while continuous probabilities from a downstream classifier (e.g., Random Forest or LR10) are aggregated by mean across runs. The resulting risk score routes the case to standard scheduling or expedited human review. On the held-out test set, the Random Forest pipeline reaches 85.9\% AUC (95\% CI 77.7--92.2), statistically equivalent to the same classifier on clinician-scored inputs ($p>0.3$). Critically, the per-feature scores remain inspectable: a clinician reviewing a flagged case sees a human-interpretable behavioral profile rather than a black-box probability, a useful way to augment and enhance clinical decisions including generalists vs rarer specialists.

Several considerations bound the strength of these conclusions. The training corpus is a single multi-source dataset (US, curated YouTube, Bangladeshi); although source diversity helps, generalization to additional geographic, demographic, and recording contexts requires prospective validation. Four hundred fine-tuning videos is a small dataset for a model of Gemini~2.5~Pro's parameter count;\citep{comanici2025gemini} the single regressed feature (head nod/shake) is consistent with mild domain overfitting, and we expect larger fine-tuning corpora and broader behavioral coverage to reduce residual feature-level regressions. The 99-video held-out test set, although independently withheld and stratified, is modest, and confidence intervals on per-feature $\kappa$ for low-prevalence items remain wide. Finally, Gemini~2.5~Pro is a closed proprietary model accessed through a managed cloud interface, which limits independent replication and exposes any deployment to silent vendor-side updates; further studies should benchmark open-weight multimodal models on the same pipeline and explore ensembling across model families to increase robustness. Within these bounds, the present results establish that supervised fine-tuning of a multimodal LLM can deliver clinician-grade behavioral feature scoring for autism screening from short home videos, and that the architecturally appropriate deployment of such systems is as a feature extractor feeding a validated supervised classifier rather than a stand-alone diagnostic oracle. 

Looking forward, systems built on this paradigm could meaningfully augment scarce human diagnostic capacity at population scale, particularly in low- and middle-income countries and in rural communities where specialty assessment is limited or unavailable today, shortening the path from caregiver concern to early intervention in precisely the populations that currently wait the longest.

\section{Methods}

\begin{figure*}[t]
\centering
\includegraphics[width=\textwidth]{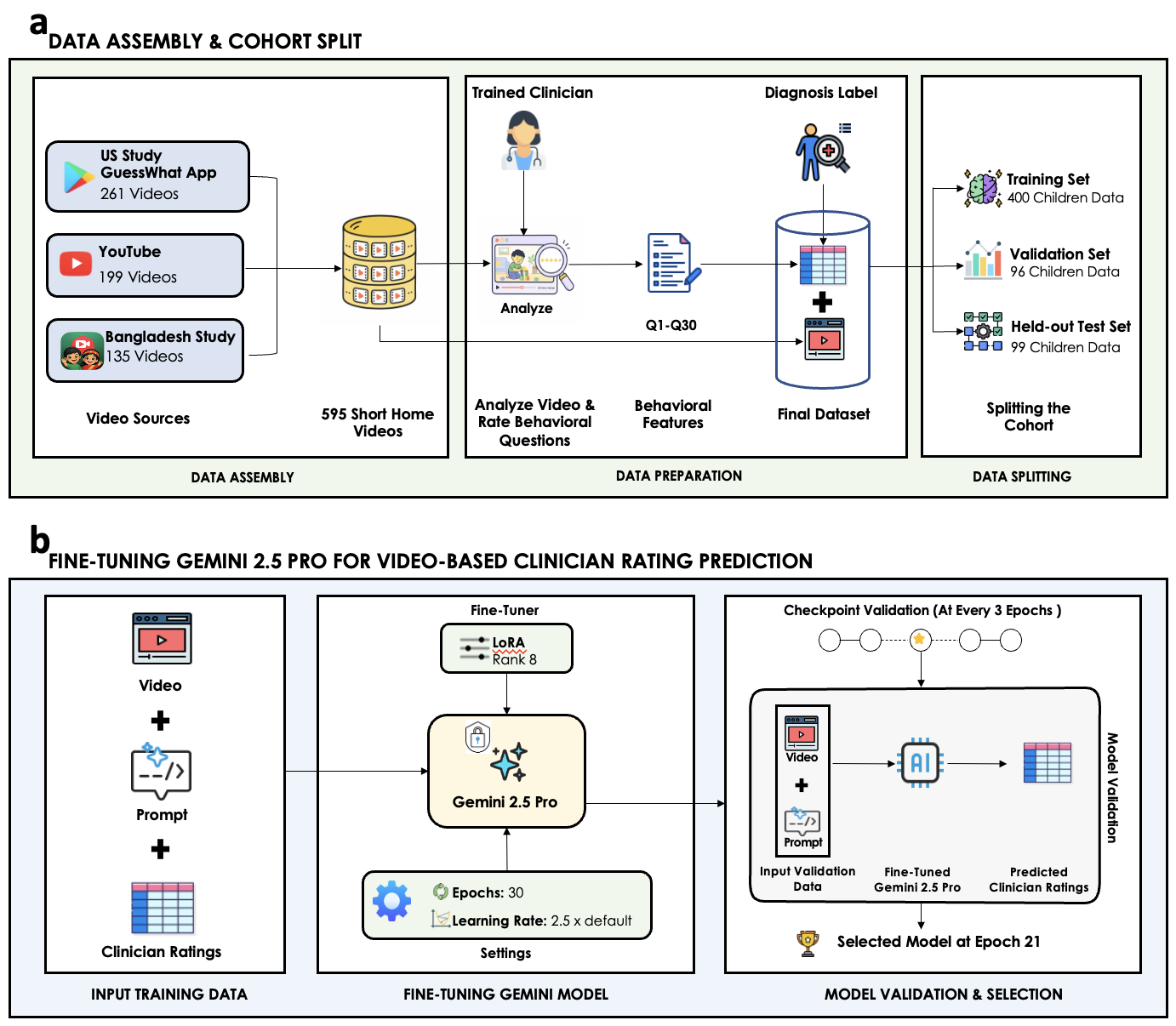}
\caption{\textbf{Study design: data curation, clinician annotation, and supervised fine-tuning.} \textit{Top:} 595 short home videos of children were assembled from three complementary sources---the GuessWhat US cohort ($n=261$), curated public YouTube clips ($n=199$), and a Bangladesh research study ($n=135$)---selected to span phenotypic, demographic, and recording-context diversity. Qualified clinicians viewed each video and rated 30 behavioral items (Q1--Q30) on categorical and ordinal scales; the separate global ASD-vs-neurotypical (NT) diagnostic label was obtained from each child's study metadata rather than assigned by the clinician. The corpus was stratified by source family and diagnosis and split into training ($n=400$), validation ($n=96$), and held-out test ($n=99$) videos, disjoint at the child level. \textit{Bottom:} each fine-tuning example pairs a video and a fixed behavioral-scoring prompt (input) with the clinician's Q1--Q30 rating vector (supervised target); the diagnostic label was withheld from training. On Stanford's HIPAA-compliant Google Cloud (PHI Secure),Gemini~2.5~Pro was adapted on the 400 training videos using low-rank adaptation (LoRA, adapter rank~8) for 30 epochs at a $2.5\times$ learning-rate multiplier. Ten checkpoints saved every three epochs were each evaluated on the 96-video validation set, and the epoch-21 checkpoint---which jointly satisfied the four-metric validation framework (Fig.~\ref{fig:val_metrics})---was selected for all downstream evaluation. [Icons used in this figure were made by awicon, meaicon, berkahicon, Parzival’ 1997, Flat Icons, Freepik, Vectors Market, nawicon, FACH, and Amazona Adorada from www.flaticon.com, used under the Flaticon Free License. The figure also includes icons created with the assistance of OpenAI’s ChatGPT (chat.openai.com) using generative AI. Use of these assets complies with OpenAI’s terms, which permit reuse for publication and commercial purposes.]}
\label{fig:overview}
\end{figure*}

\begin{figure*}[tp]
\centering
\includegraphics[width=\textwidth]{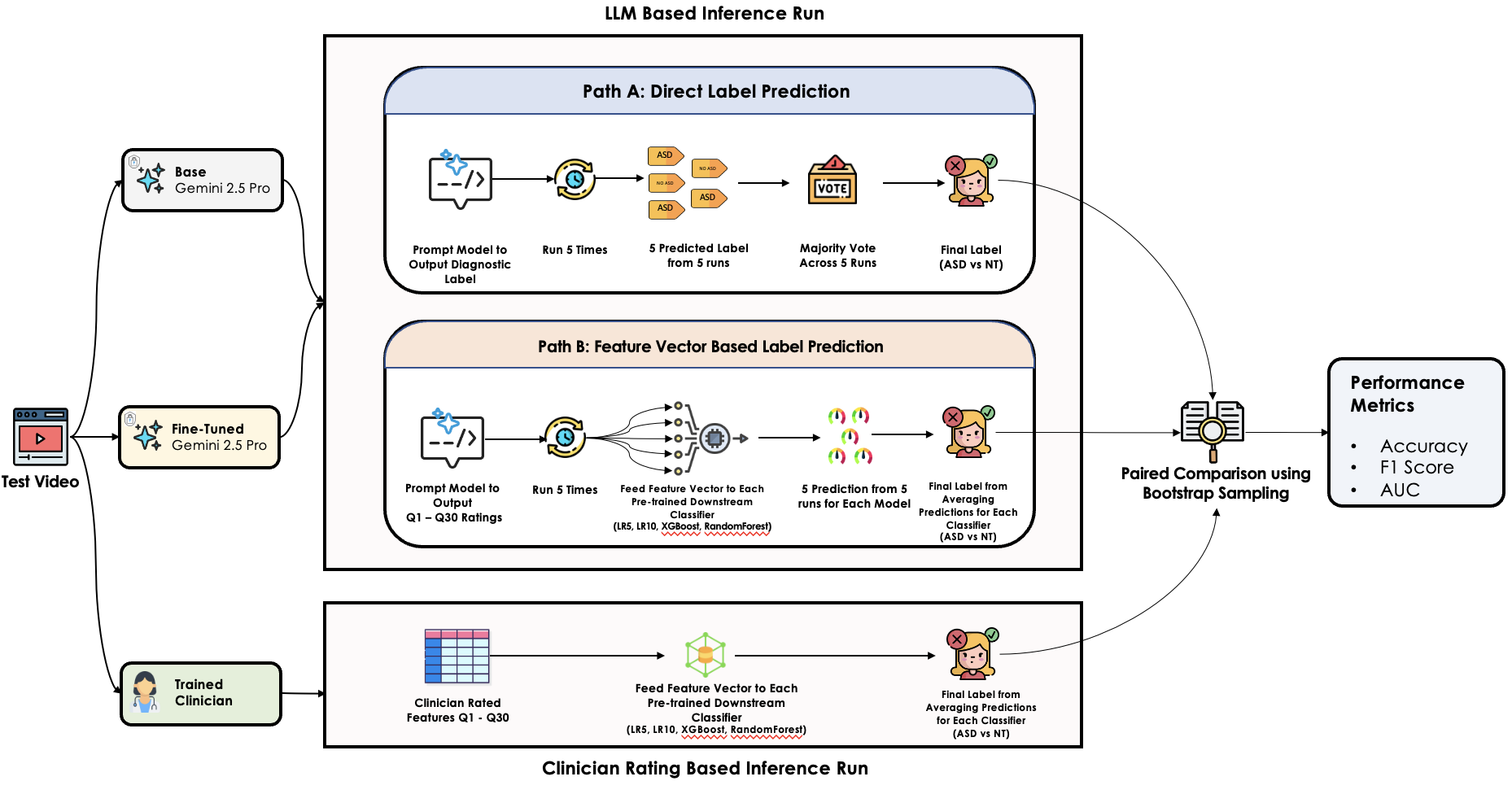}
\caption{\textbf{Evaluation design: diagnostic pathways and feature sources.} Each diagnostic call on the 99-video held-out test set is produced in one of two ways. \textbf{Path~A (direct label prediction):} the model is prompted to output the binary ASD-vs-NT label directly; five independent inference runs are aggregated by majority vote. \textbf{Path~B (feature-vector-based prediction):} the model is prompted to output the 30 behavioral ratings (Q1--Q30), and each run's feature vector is passed to a panel of pre-trained downstream classifiers (LR5, LR10, XGBoost, Random Forest); per-classifier probabilities are averaged across the five runs before the operating-threshold decision. The clinician pathway (bottom) applies the same downstream classifiers to clinician-scored Q1--Q30 features and provides the ground truth reference. Base and fine-tuned Gemini~2.5~Pro are both evaluated through Paths~A and~B; the figure pairs each model with the pathway that defines its principal result. [Icons used in this figure were made by Uniconlabs, nangicon, riajulislam, Freepik, Parzival’ 1997, ChilliColor, nawicon, FACH, and Flat Icons from www.flaticon.com, used under the Flaticon Free License.]}
\label{fig:eval}
\end{figure*}
\subsection{Data Collection}

All videos were collected under an approved study protocol reviewed and approved by the Stanford Institutional Review Board (IRB) in accordance with Stanford, California, United States, and~international research guidelines, including the Declaration of Helsinki, prior to the initiation of any study procedures.

GuessWhat videos  were collected through the mobile game GuessWhat~\cite{kalantarian2019guess}, designed to facilitate structured, charades-style interactions between parent and child.  Informed consent was obtained from parents/legal guardians through the app before study participation. 

Videos from Bangladesh were collected at Dhaka Shishu Children’s Hospital (DSH) and Square Hospital (SH) in Bangladesh under institutional review board (IRB)-approved protocols at those hospitals and at Stanford University. Videos were collected by clinical staff at DSH and SH following informed consent by a parent/legal guardian. Brief videos (2-5 minutes) were recorded during evaluation of the children who presented to the DSH and SH Child Development Centers with neurodevelopmental concerns.~\cite{tariq2019detecting}  

\subsection{Cohort and dataset splits}

The corpus comprised 595 short home videos of children, each annotated by trained clinicians on a 30-item behavioral questionnaire (Q1--Q30) and a separate global ASD-vs-NT diagnostic label. Videos were drawn from three source families: (i)~a US cohort centred on the GuessWhat study\citep{kalantarian2019guess} (261 videos: 208 ASD, 49 NT, 4 Other); (ii)~curated public YouTube clips (199 videos: 67 ASD, 50 NT, 82 Other); and (iii)~a Bangladeshi research study (135 videos: 47 ASD, 30 NT, 58 Other). Diagnoses were obtained by linking video metadata to a child-level registry; raw labels were normalised to ASD, NT, or Other. Videos were pre-filtered to $\leq$100\,MB; typical clip duration was approximately 89~seconds (file sizes 5--21\,MB).

Splits were generated with a fixed seed (\texttt{numpy.random.seed(42)}) and stratified by source-family $\times$ diagnosis to preserve both source and class balance, yielding 400 training videos, 96 validation videos, and 99 held-out test videos (per-split breakdown in Table~\ref{tab:cohort}). Disjointness was verified programmatically at the child level. ASD prevalence was matched between training (55.0\%) and validation (55.2\%); the held-out test set was constructed to be balanced (49 ASD, 50 NT).

\subsection{Behavioral scoring instrument}

The model was tasked with rating each video on 30 behavioral features (Q1--Q30, used in prior work\citep{azizian2025multimodal}) and producing a separate binary global ASD-vs-NT diagnostic label. The 30 items cover the diagnostically central observable domains, including language and speech, social gaze and nonverbal communication, social affect and initiation, play quality, and sensory and repetitive behaviors. Most items use 4-point ordinal scales (0--3); a small number use 3- or 5-point scales. A universal `\texttt{8}=N/A' code indicates that the relevant behavior was genuinely unobservable in the video. Only Q1--Q30 were used as fine-tuning targets; the diagnostic label was held out from training.

\subsection{Fine-tuning procedure}

The fine-tuning task is structured as supervised next-token prediction with a clinician-provided target. For each video clip, a clinician produced an ordinal rating for each of Q1--Q30; the model is then trained to reproduce that rating vector when given the video and the behavioral-scoring prompt as input (Fig.~\ref{fig:overview}). Each training example was encoded as a multimodal JSONL entry: the user turn supplied the video file (Google Cloud Storage URI, MIME type \texttt{video/mp4}) and the behavioral-scoring prompt; the model turn supplied the target as a JSON object mapping each item to its clinician rating. The final training set comprised 400 entries; the validation set 96 entries.

\begin{figure}[t]
\centering
\includegraphics[width=\linewidth]{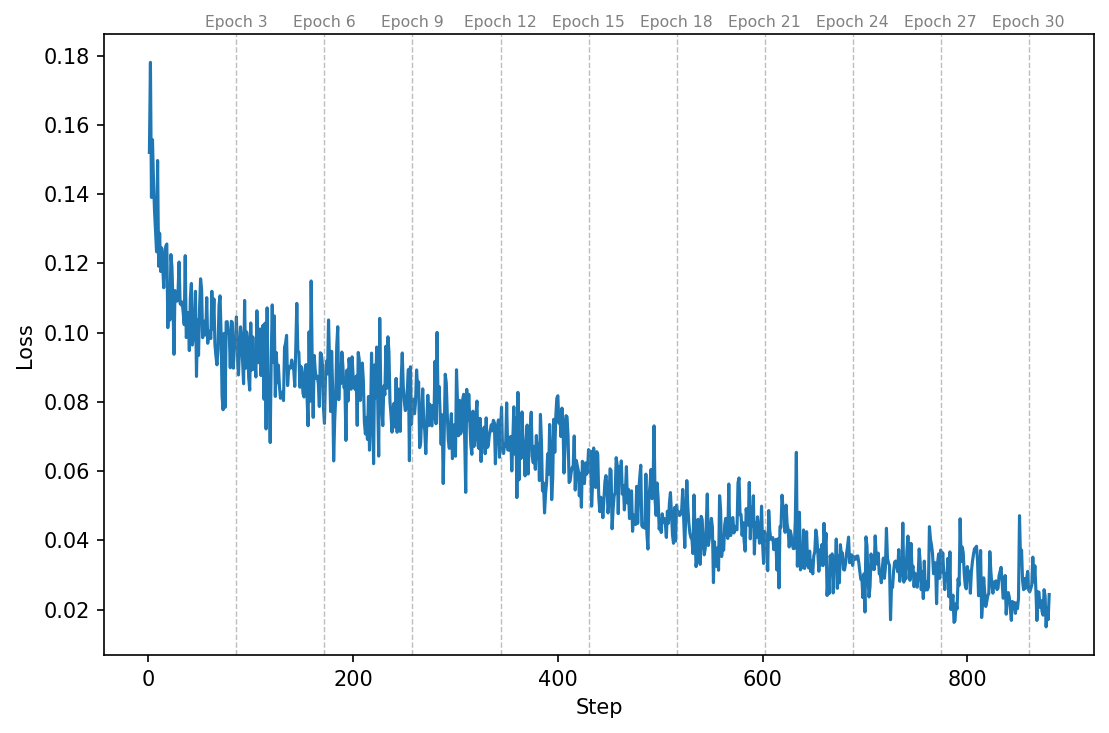}
\caption{\textbf{Training loss over 30 epochs of supervised fine-tuning.} Per-step training loss (Eq.~\ref{eq:sft_loss}) over 880 gradient steps ($\approx$29.3 steps/epoch). Vertical dashed lines mark the 10 saved checkpoints (every 3 epochs). Loss decreases steadily from $\approx$0.18 at initialisation to $\approx$0.02--0.04 at epoch~30, with no late-stage divergence; epoch~21 was selected as the final checkpoint based on the four-metric validation framework (Fig.~\ref{fig:val_metrics}).}
\label{fig:loss}
\end{figure}

Fine-tuning was performed on Stanford's HIPAA-compliant Google Cloud (PHI Secure) using Vertex AI Supervised Fine-Tuning (\texttt{vertexai.preview.tuning.sft.train()}) on \texttt{gemini-2.5-pro}.\citep{comanici2025gemini} Vertex AI SFT applies low-rank adaptation (LoRA) by default; we used adapter rank 8, a learning-rate multiplier of $2.5\times$ the Vertex default, and trained for 30 epochs over 400 examples (880 gradient steps, $\approx$29.3 steps per epoch). The training objective is the standard autoregressive (causal) language-modelling loss over the target token sequence:
\begin{equation}
\mathcal{L}_{\text{SFT}}(\theta) = -\,\mathbb{E}_{(\mathbf{x},\mathbf{y})\sim\mathcal{D}_{\text{train}}}\!\!\sum_{t=1}^{|\mathbf{y}|}\log p_{\theta}(y_t \mid y_{<t}, \mathbf{x}),
\label{eq:sft_loss}
\end{equation}
where $\mathbf{x}$ is the multimodal input (video and prompt), $\mathbf{y}=(y_1,\ldots,y_{|\mathbf{y}|})$ is the JSON target sequence containing the 30 clinician ratings, and $\theta$ are the LoRA adapter parameters. Only the 30 behavioral feature token positions contribute to $\mathcal{L}_{\text{SFT}}$; the diagnostic label was held out from training so that no diagnostic supervision could leak into the learned feature representation. Training loss decreased monotonically over the 30 epochs (Fig.~\ref{fig:loss}), from $\approx$0.18 at initialisation to $\approx$0.02--0.04 by epoch~30, with no late-stage divergence.

\subsection{Checkpoint selection}

Vertex AI saved 10 checkpoints (every 3 epochs); each was deployed as a Vertex AI endpoint and evaluated on the 96-video validation set at temperature~0. Because the model may legitimately abstain when a behavior is unobservable, we did not reduce agreement to a single accuracy or $\kappa$ score. Each (model, clinician) response pair was partitioned into one of four mutually exclusive regions (Fig.~\ref{fig:ratability_regions}): $N_1$ (model abstains, clinician rates), $N_2$ (both abstain), $N_3$ (clinician abstains, model rates), and $N_4$ (both rate). Four complementary metrics were derived from these regions:
\begin{itemize}\setlength{\itemsep}{0.15em}
    \item \textbf{Quadratic weighted Cohen's $\kappa_w$} on $N_4$ (primary metric for ordinal agreement; quadratic weighting penalises larger disagreements more heavily).
    \item \textbf{Ratability Agreement (RA)}: $\mathrm{RA}=(N_2+N_4)/(N_1+N_2+N_3+N_4)$, the fraction of items where model and clinician agree on whether to rate at all.
    \item \textbf{Bias Index (BI)}: $\mathrm{BI}=(N_3-N_1)/(N_1+N_2+N_3+N_4)$, where positive values indicate the model rates when the clinician abstains and negative values indicate the opposite. A value near zero indicates no systematic over- or under-abstention.
    \item \textbf{False Abstention Rate (FAR)}: $\mathrm{FAR}=N_1/(N_1+N_2+N_3+N_4)$, the fraction of items for which the model incorrectly abstained.
\end{itemize}

\IfFileExists{figures/ratability_regions.png}{%
\begin{figure}[t]
\centering
\includegraphics[width=\linewidth]{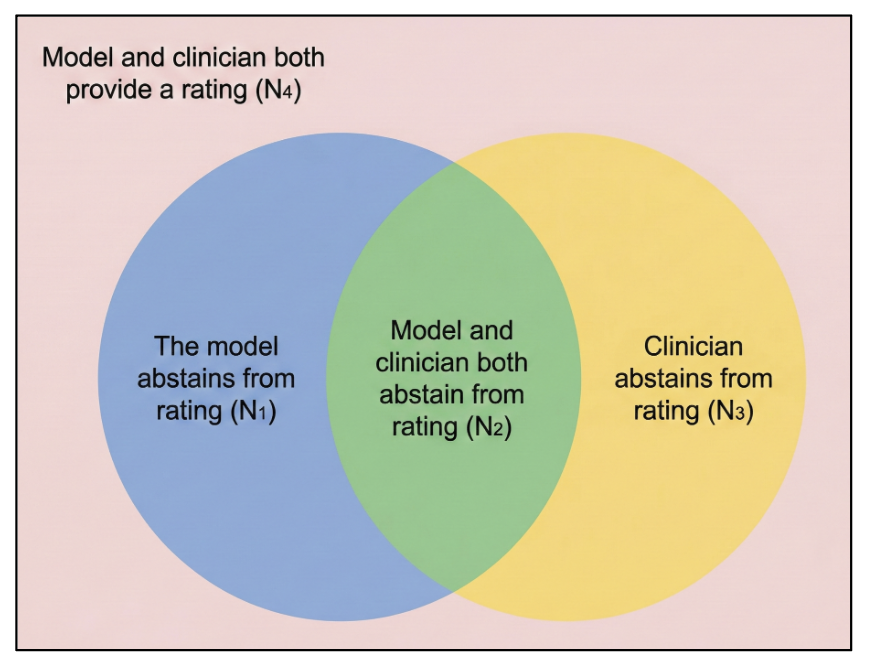}
\caption{\textbf{Four-region partition of (model, clinician) response pairs underlying the validation framework.} For every (video, item) pair both the clinician and the model independently choose between providing an ordinal rating and abstaining (`\texttt{8}=N/A'). This induces four mutually exclusive regions: $N_1$ (model abstains, clinician rates), $N_2$ (both abstain), $N_3$ (clinician abstains, model rates), and $N_4$ (both rate). $\kappa_w$ is computed on $N_4$; RA uses $N_2{+}N_4$; BI uses $N_3{-}N_1$; FAR uses $N_1$ (each normalised by the total $N_1{+}N_2{+}N_3{+}N_4$).}
\label{fig:ratability_regions}
\end{figure}
}{}

\begin{figure*}[t]
\centering
\begin{minipage}{0.49\linewidth}\centering
  \includegraphics[width=\linewidth]{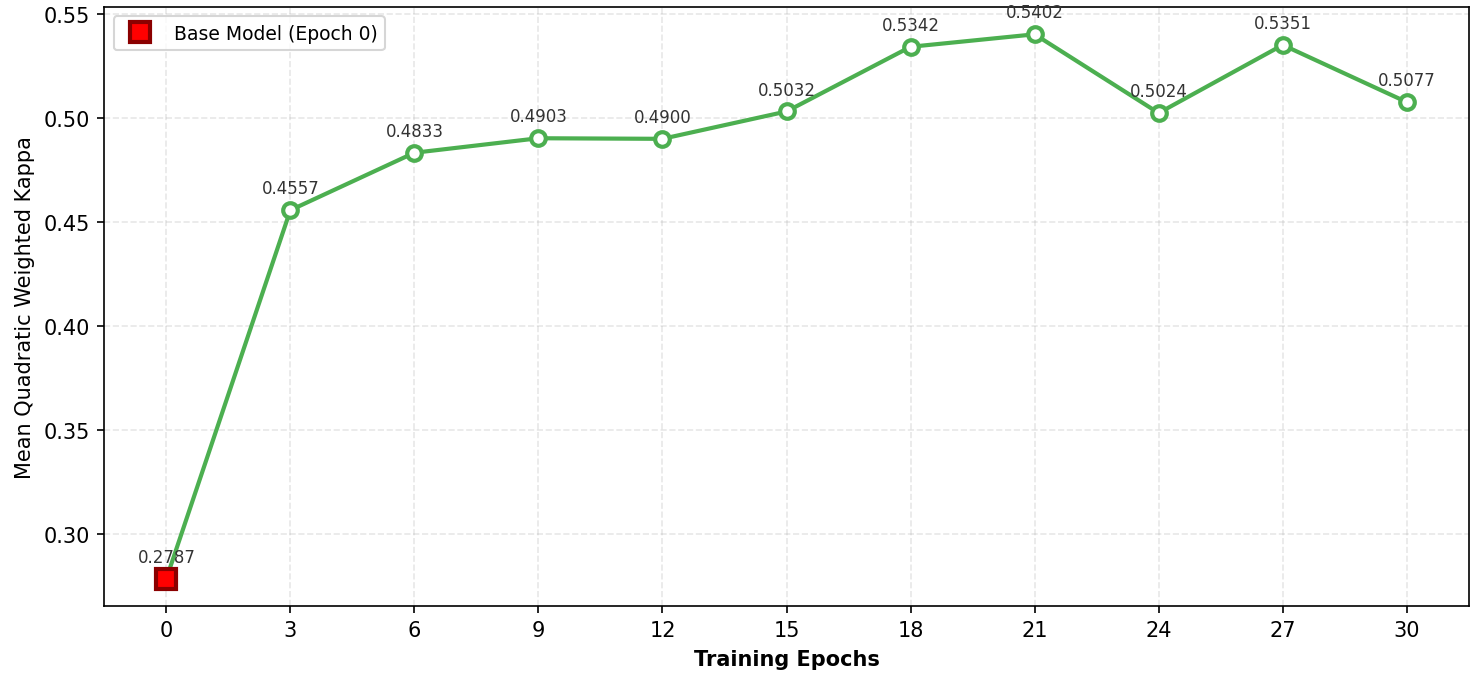}\\
  \footnotesize (a) Weighted Cohen's $\kappa_w$
\end{minipage}\hfill
\begin{minipage}{0.49\linewidth}\centering
  \includegraphics[width=\linewidth]{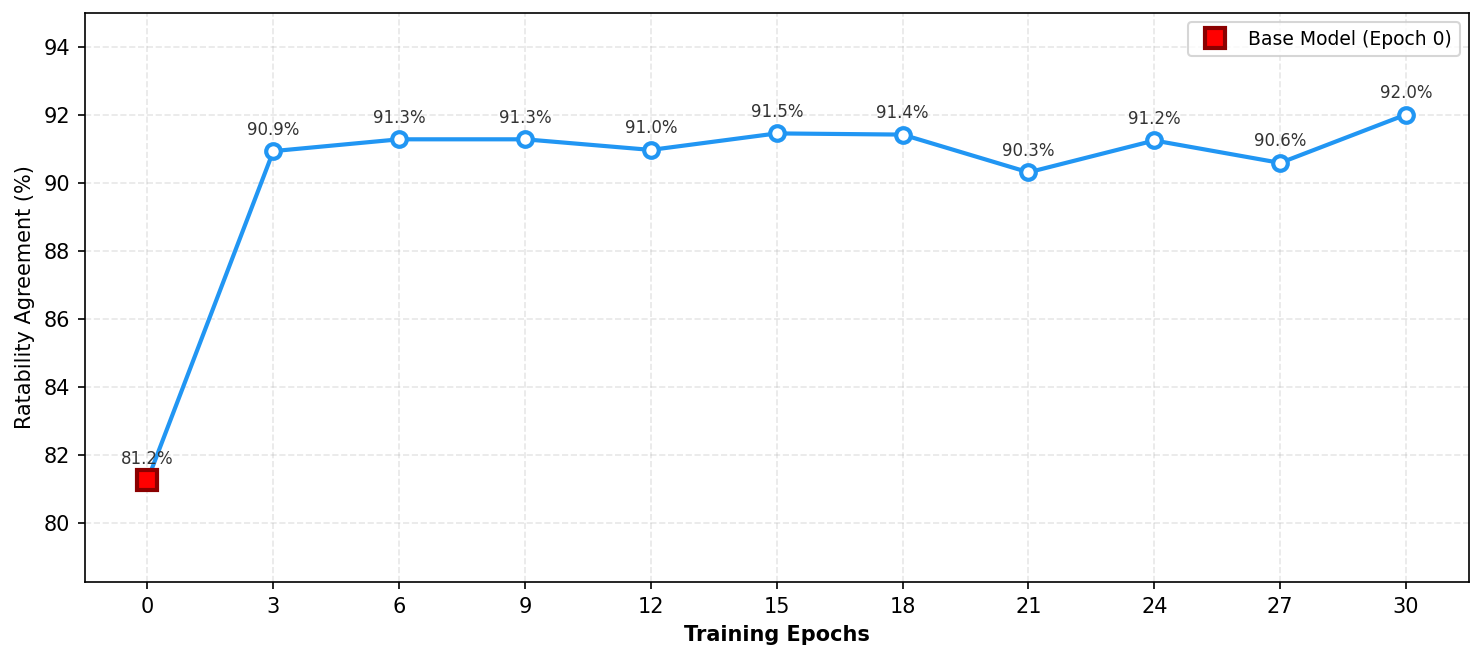}\\
  \footnotesize (b) Ratability Agreement (\%)
\end{minipage}

\vspace{1em}

\begin{minipage}{0.49\linewidth}\centering
  \includegraphics[width=\linewidth]{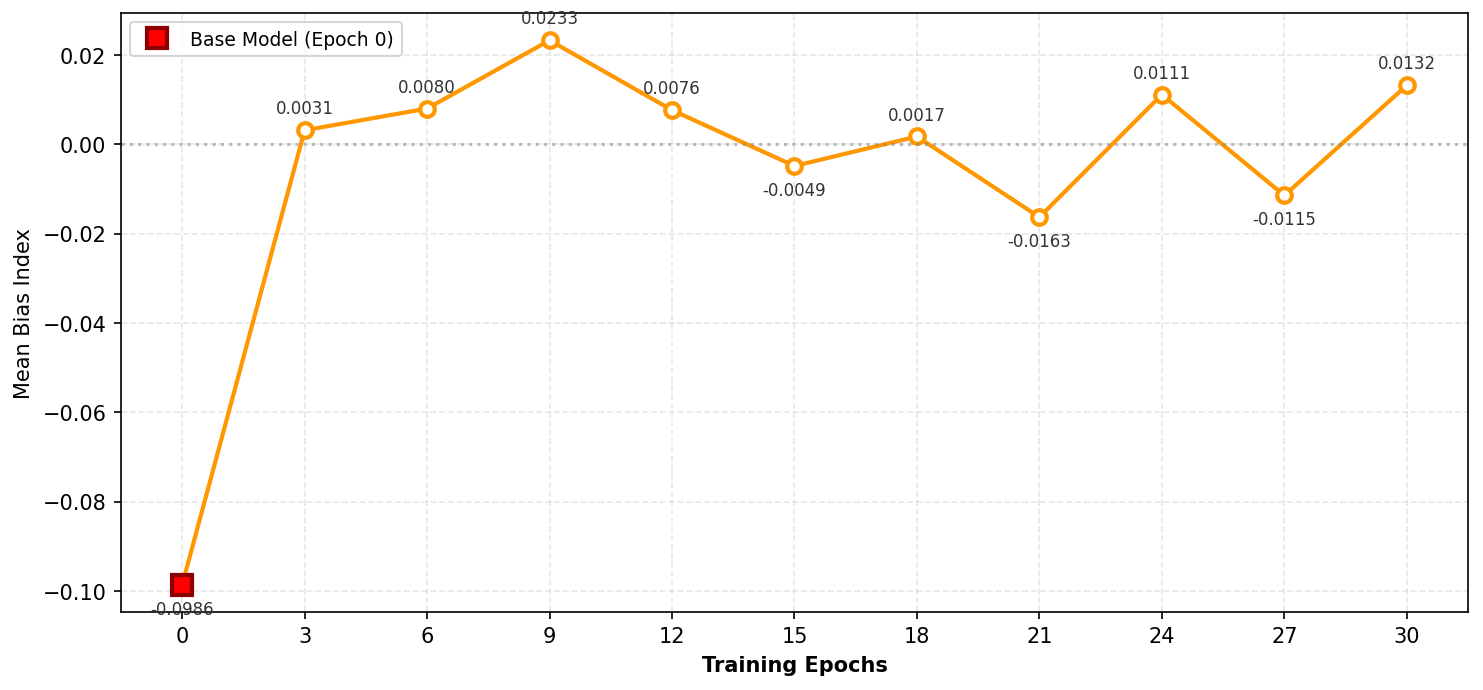}\\
  \footnotesize (c) Bias Index (signed)
\end{minipage}\hfill
\begin{minipage}{0.49\linewidth}\centering
  \includegraphics[width=\linewidth]{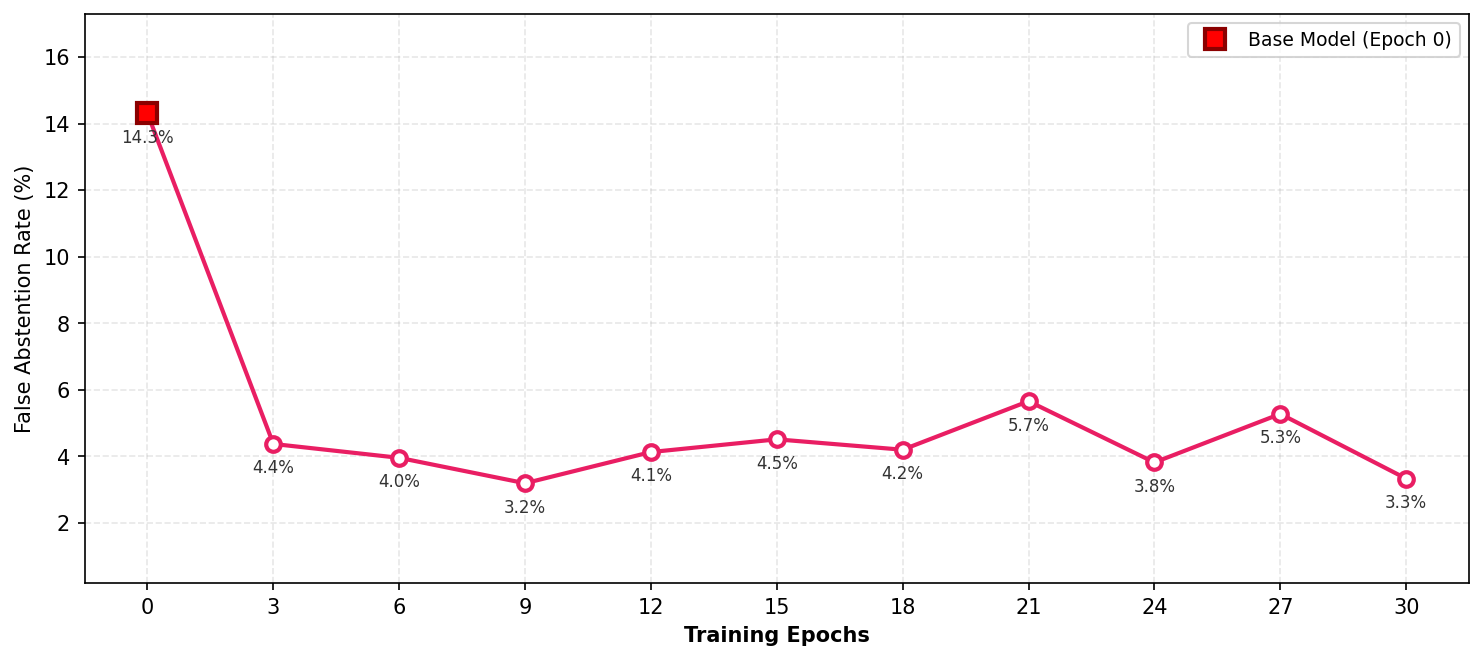}\\
  \footnotesize (d) False Abstention Rate (\%)
\end{minipage}
\caption{\textbf{Four-metric validation curves used for checkpoint selection.} Each panel plots the metric on the 96-video validation set as a function of training epoch (epoch~0 = base model, plotted as a red square). \textbf{(a)}~Weighted Cohen's $\kappa_w$ rises rapidly from 0.28 at epoch~0 to 0.46 by epoch~3 and continues to improve more slowly, peaking at 0.540 at epoch~21 (red circle). \textbf{(b)}~Ratability Agreement saturates near 91\% by epoch~3 and remains stable. \textbf{(c)}~Bias Index of the base model is large and negative ($-0.099$) but immediately crosses zero on fine-tuning and stays within $\pm0.02$; epoch~21 has the lowest absolute BI ($-0.016$). \textbf{(d)}~False Abstention Rate drops from 14.3\% at base to a stable $\approx$3--6\% post fine-tuning. The four panels jointly motivate the selection of epoch~21 as the operating checkpoint.}
\label{fig:val_metrics}
\end{figure*}

We tracked all four metrics as a function of training epoch on the held-out validation set (Fig.~\ref{fig:val_metrics}). $\kappa_w$ rose sharply over the first 3 epochs (0.28 to 0.46), then improved more slowly to a maximum of 0.540 at epoch~21; Ratability Agreement saturated near 91\% by epoch~3 and remained stable thereafter; the Bias Index, large and negative for the base model ($-0.099$), crossed zero immediately upon fine-tuning and stabilised within $\pm0.02$ of zero across all checkpoints; and the False Abstention Rate fell from 14.3\% (base) to a stable $\approx$3--6\% by epoch~3. Epoch~21 was selected for all downstream evaluation because it simultaneously achieved the highest validation $\kappa_w$ (0.540), Bias Index very close to zero, and Ratability Agreement and False Abstention Rate at their post-fine-tuning saturation values, satisfying all four metrics at once.

\subsection{Inference and ensembling}

All test-set evaluation used the 99-video held-out set (49 ASD, 50 NT) and compared two model conditions, base Gemini~2.5~Pro and the fine-tuned epoch~21 checkpoint. Each model was queried five independent times per video at temperature~0.5, yielding 990 inferences per condition. For the direct ASD-vs-NT diagnostic call, the five binary outputs were aggregated by majority vote. For classifier-assisted pathways, the downstream classifier was applied to each of the five feature vectors, and the resulting probability outputs were averaged across runs (mean of five probability outputs) before the operating-threshold decision. The model was prompted to provide an N/A response for any item it judged unobservable; N/A responses were excluded from $\kappa$ calculations as in the validation framework above.

\subsection{Downstream classifiers}

To evaluate whether LLM-predicted behavioral feature scores encode sufficient signal for ASD diagnosis, the predicted Q1--Q30 scores were passed to a panel of classifiers validated in previous home-video screening studies.\citep{tariq2018mobile,kosmicki2015searching,levy2017sparsifying} The panel comprised two pre-trained logistic-regression classifiers (LR5 on five ADOS Module~2 features for children with limited speech; LR10 on ten ADOS Module~3 features for verbally fluent children), and two non-linear classifiers (XGBoost and Random Forest) trained on a separate crowdsourced behavioral-rating corpus of 7{,}037 records (after removing the 99 test-set children at the child level to prevent leakage). We additionally evaluated direct LLM diagnosis (the model's diagnostic-label ensemble call) as a fifth pathway. A fixed correspondence table mapped behavioral items to the inputs expected by each classifier. Each pathway was scored under three input conditions: base Gemini, clinician annotations (ground truth), and fine-tuned Gemini (Fig.~\ref{fig:eval}).

\subsection{Statistical analysis}

Per-feature $\kappa$ confidence intervals were computed by paired bootstrap resampling at the (video, feature) pair level (10{,}000 resamples). For classifier metrics, paired bootstrap tests on the 99-video test set (10{,}000 resamples) compared each pair of input conditions (base vs clinician, clinician vs fine-tuned, base vs fine-tuned) within each pathway and metric (accuracy, AUC, F1). All reported $p$-values are two-sided. Q31 sensitivity confidence intervals were computed by Clopper-Pearson exact method on $n=49$ ASD cases. Significance markers in figures: $\dot{}~p<0.10$, $*~p<0.05$, $**~p<0.01$, $***~p<0.001$.

\subsection{Use of AI-assisted tools in manuscript preparation}

During manuscript preparation, the authors used Claude Opus 4.8 to assist with drafting and editing text, refining manuscript structure, formatting the presentation of results, and improving clarity and coherence. The authors reviewed and edited the output and take full responsibility for the content of the manuscript.

\section*{Data availability}
The data presented in this article may be made available upon request and upon entering a Data Use Agreement with the corresponding institutions to gain access to a limited dataset.

\section*{Code availability}
The underlying code used for data preprocessing, model-output parsing, statistical analysis, and figure generation is not publicly available because it contains study-specific data-handling logic, access-controlled file paths, and cloud-inference configuration. The code will be made available to editors and reviewers upon request for the purposes of peer review.

\section*{Author contributions}
M.H. and P.A. conceived the study and designed the fine-tuning and evaluation methodology. M.H. and P.A. implemented the fine-tuning pipeline, ran the model training and inference experiments, performed the statistical analyses, and generated the figures. A.Kl. contributed to methodology, validation, and interpretation of the results. K.N. and S.S. contributed to the development and evaluation of the machine-learning classifiers. Z.N.T. contributed to figure generation and visualization. K.D. contributed to data curation. Y.Q., A.Ka., and S.N. contributed to manuscript review and revision. P.Y.W. and D.P.W. supervised the project and provided clinical and scientific guidance. M.H. wrote the original draft. All authors reviewed and edited the manuscript. M.H. and P.A. contributed equally. All authors read and approved the final manuscript.

\section*{Acknowledgements}
This work was funded by the National Institutes of Health (R01LM014342, R01LM013364 (Wall), the Stanford Medicine Center for Digital Health, Stanford Wu Tsai Neurosciences Institute, Stanford University Human-Centered Artificial Intelligence, and the Islamic Development Bank Transform Fund. PW acknowledges support from award DP2-EB035858 from the NIH. The content is solely the responsibility of the authors and does not necessarily represent the official views of the funding organizations. The study was conducted in accordance with the Declaration of Helsinki, and approved by the Institutional Review Boards at Stanford University (protocol 39562), Square Hospital, and Dhaka Shishu Hospital. Informed consent was obtained from all subjects involved in the study prior to any research procedures taking place. We thank the families who contributed videos and the clinical raters who provided the ground-truth behavioral annotations on which this work depends, and Google for research credits supporting the Vertex AI fine-tuning experiments.

\section*{Competing interests}
The authors declare no competing financial or non-financial interests.

\bibliography{references}

\end{document}